\documentclass[letterpaper, 10 pt, conference]{ieeeconf}  

\IEEEoverridecommandlockouts                              

\usepackage[backend=biber,
            hyperref=true,
            url=false,
            isbn=false,
            doi=false,
            backref=false,
            style=ieee,
            citestyle=numeric-comp,
            sorting=nyt,
            block=none]{biblatex}

\usepackage{flushend}
\usepackage{balance}

\usepackage{multicol}
\usepackage[bookmarks=true]{hyperref}
\usepackage[algo2e, vlined, ruled, linesnumbered]{algorithm2e}
\usepackage{color}
\usepackage{algorithm}
\usepackage{algpseudocode}
\usepackage{graphicx}
\usepackage{subcaption}
\usepackage{amsmath}
\usepackage{amsfonts}
\usepackage{booktabs}

\title{\LARGE \bf
Generalizable Task Planning through Representation Pretraining
}

\author{Chen Wang$^{1}$, Danfei Xu$^{2}$, Li Fei-Fei$^{1}$
\thanks{$^{1}$ Stanford Vision and Learning Lab $^{2}$ NVIDIA}
}

\begin{document}

\maketitle
\thispagestyle{empty}
\pagestyle{empty}

\begin{abstract}

The ability to plan for multi-step manipulation tasks in unseen situations is crucial for future home robots. But collecting sufficient experience data for end-to-end learning is often infeasible in the real world, as deploying robots in many environments can be prohibitively expensive. On the other hand, large-scale scene understanding datasets contain diverse and rich semantic and geometric information. But how to leverage such information for manipulation remains an open problem.
In this paper, we propose a learning-to-plan method that can generalize to new object instances by leveraging object-level representations extracted from a synthetic scene understanding dataset. We evaluate our method with a suite of challenging multi-step manipulation tasks inspired by household activities~\cite{srivastava2022behavior} and show that our model achieves measurably better success rate than state-of-the-art end-to-end approaches. Additional information can be found at: \url{https://sites.google.com/view/gentp}

\end{abstract}

\section{Introduction}

Planning for everyday manipulation tasks in home environments is challenging for multiple reasons. The problem not only requires searching in high-dimensional, non-convex spaces over long time horizons but also poses major \emph{representation challenges}. For example, consider the task of cleaning a dining table with a towel by soaking the towel first. The planner needs to represent both geometric (e.g. location of the sink) and semantic (e.g., a towel is soaked) states as well as how actions might affect these states. While research fields such as Task and Motion Planning (TAMP) have made significant progress in solving these tasks efficiently~\cite{lozano2014constraint, garrett2021integrated, garrett2020pddlstream, toussaint2015logic, toussaint2018differentiable}, most of the approaches rely on carefully-chosen abstract representations and analytically-defined transition models, both of which are often domain- or even task-specific. Thus for a home robot to be effective, it is vital to equip its planner with planning representations that can generalize to a wide range of tasks and environments.

Fortunately, recent progress in deep learning has shown that it is possible to extract generalizable representations from raw perception data such as images. A particular relevant thread is in visual reasoning, where works~\cite{mao2019neuro,radford2021learning,yuan2021sornet} have shown that implicit object-centric representations learned through large-scale scene understanding tasks such as predicting spatial relationships can be transferred to new tasks and generalize to unseen objects. However, there are two open challenges when applying similar recipes to learning planning representations. First of all, planning with manipulation skills requires a certain level of \emph{representation granularity}. For example, modeling the effects of a \emph{placing} skill requires reasoning about the spatial locations and the shapes of an object and a target surface. Representations derived from just learning abstract concepts such as \texttt{on-top} may not be sufficient for planning. What pretraining strategies can allow us to extract suitable planning representations? Second, planning systems such as TAMP assume explicit states (e.g., object poses) and world models (e.g., kinematic transition models). How can we develop a method that can plan with learned implicit representations?

In this paper, we aim to develop a learning-to-plan method that can generalize to new objects and task goals through addressing these two problems. On a high level, we propose a two-stage approach. The first stage extracts object-level embeddings from a dataset of raw RGB observations using a suite of auxiliary pretraining objectives. The goal is to find invariant features that generalize to new objects without additional finetuning. The objectives include learning high-level semantic attributes and tasks that require more fine-grained geometric understanding such as pixel-level segmentation. We empirically show that such hybrid learning objectives are crucial for the subsequent stage of learning to plan with parameterized manipulation skills. Moreover, the learned representations do not pertain to specific planning problems and thus can be reused across tasks. 

\begin{figure}[t]
	\centering
    \includegraphics[width=1.0\linewidth]{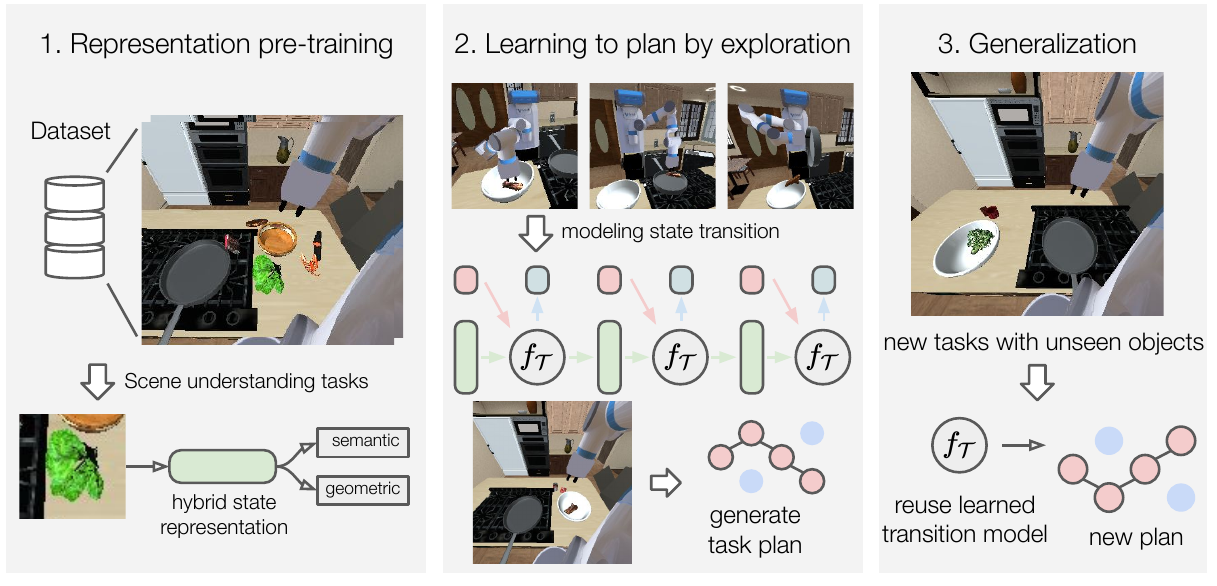}
	\caption{\textbf{Overview}: We propose a two-stage approach to learn a generalizable task planner. (1) The first stage learns object-centric representations from a large-scale scene understanding dataset through hybrid semantic-symbolic objectives. (2) Based on the pre-trained representation, the second stage trains an environment dynamics model $f_{\mathcal{T}}$ for search-based task planning. (3) We show that the task planner inherits the generalization capabilities of the learned representation and can successfully plan in new settings with unseen objects.}
	\label{fig:pullfig}
\end{figure}

The second stage is a novel planning framework that builds a dynamics model of the environment based on the learned object representations. We show that our search-based planner can plan for different tasks with a single learned transition model. More importantly, the learned object representations allow the planner to generalize to new objects with no retraining or finetuning. At the same time, learning to plan with unconstrained, continuously-parameterized skills still poses a major technical challenge. To this front, we propose a novel dynamics model that \emph{grounds} the skill parameters in the learned planning space, allowing the model to directly associate the effects of skill with the implicit representation of a target object. We show that the model greatly enhances the robot's ability to solve tasks that require more precise skill parameter selection.

We evaluate our method with three manipulation tasks of varying difficulties. The \textit{Rearrangment} task is to rearrange two (possibly unseen) objects that are placed on two tables. In the \textit{Cleaning} task, the robot needs to first soak a towel in a sink and use the soaked towel to clean a microwave oven, testing the planner's ability to represent both geometric and semantic states. Finally, in the most challenging \textit{Cooking} task, the robot is asked to plan for cooking unseen food objects with a pan and serve the cooked food in a bowl. We empirically demonstrate our method's ability to plan for complex multi-step manipulation tasks and generalize to new objects and task goals.

\section{Related works}

\subsection{Task and Motion Planning}

Task and Motion Planning (TAMP) methods are effective at solving multi-step robot manipulation tasks \cite{lozano2014constraint, garrett2021integrated, garrett2020pddlstream, toussaint2015logic, toussaint2018differentiable}. However, most established TAMP methods rely on analytically-defined components such as preimage functions and environment kinematics models. More recent works sought to replace these components with learned modules. Notable examples include learning to predict plan feasibilities~\cite{kaelbling2017learning, driess2020deep} and learning skill effect models~\cite{wang2021learning, liang2021search}. While these works have relaxed many assumptions of TAMP methods, the learned modules are often built on top of hand-defined representation spaces. For example, Wang \emph{et al.}~\cite{wang2021generalization} explicitly models weight changes of a container as a result of executing a pouring skill, and Liang \emph{et al.}~\cite{liang2021search} constructs skill effect model based on relative object poses. As a result, these learned modules are often limited to specific tasks or domains. Instead, we propose to learn an implicit planning space through a suite of task-agnostic representation learning objectives. We show that our learning-to-plan framework based on such representations can generalize to new tasks and object categories.

\subsection{Learning to plan}

Our method is closely related to model-based reinforcement learning~\cite{finn2017deep, hafner2019learning}. Most recent works have focused on using observation space as the state representation to build full environment models~\cite{tian2020model, wu2021example}. However, learning to make accurate predictions in, e.g., raw image space, is still challenging ~\cite{oh2017value, finn2017deep}, especially for long-horizon manipulation tasks. Instead, our approach learns a partial model of the world by first extracting features of image observation through supervised pretraining and building dynamics models based on such compact representations. 

For works that plan with partial models, many focus on predicting either reward or task-specific quantities~\cite{dosovitskiy2016learning, hafner2019learning} through end-to-end learning. As a result, it is difficult for a learned model to generalize to new tasks and settings. To address the challenge, some recent works turn to task-agnostic partial models~\cite{xu2021deep,driess2020deep,driess2021learning}. For example, Xu \emph{et al.}~\cite{xu2021deep} proposes to learn environment models based on object affordances that do not pertain to specific task goals. However, since the representations are learned together with the dynamics model in an end-to-end manner from interaction data, the learned representations are still tied to the specific settings where data is collected. Furthermore, collecting interaction data in broad and diverse settings can be prohibitively expensive in the real world. In contrast, our representation learning learns from static scene understanding dataset, which can be either generated synthetically or crowd-sources to cover diverse objects and scenarios.

\subsection{Representation learning for generalizable manipulation}

Recent research suggests that learning from diverse and large-scale dataset is a promising path towards generalizable manipulation~\cite{ebert2021bridge, jang2022bc, shridhar2021cliport}. Examples of such dataset include demonstrations across domains \cite{ebert2021bridge,jang2022bc,mandlekar2019scaling,pari2021surprising}, mixture of exploration and expert data \cite{cabi2019scaling}, and instruction-conditional data \cite{shridhar2021cliport}.

One way to leverage such datasets for robotic manipulation is learning generalizable state representations~\cite{pari2021surprising, kase2020transferable, yuan2021sornet}. For example, SORNet~\cite{yuan2021sornet} learns object embeddings that could be used to estimates the spatial relationships of unseen objects. Similarly, Cliport~\cite{shridhar2021cliport} transfers representations pretrained on a vision-language dataset~\cite{radford2021learning} to manipulation skills and show generalizations to new tasks. However, these works almost exclusively focus on single-step state estimation or learning short-horizon skills. Our work instead takes a step towards building generalization planners for multi-step manipulation problems by leveraging representations learned from large-scale datasets.

\section{Method}

\begin{figure*}[!ht]
	\centering
    \includegraphics[width=1.0\linewidth]{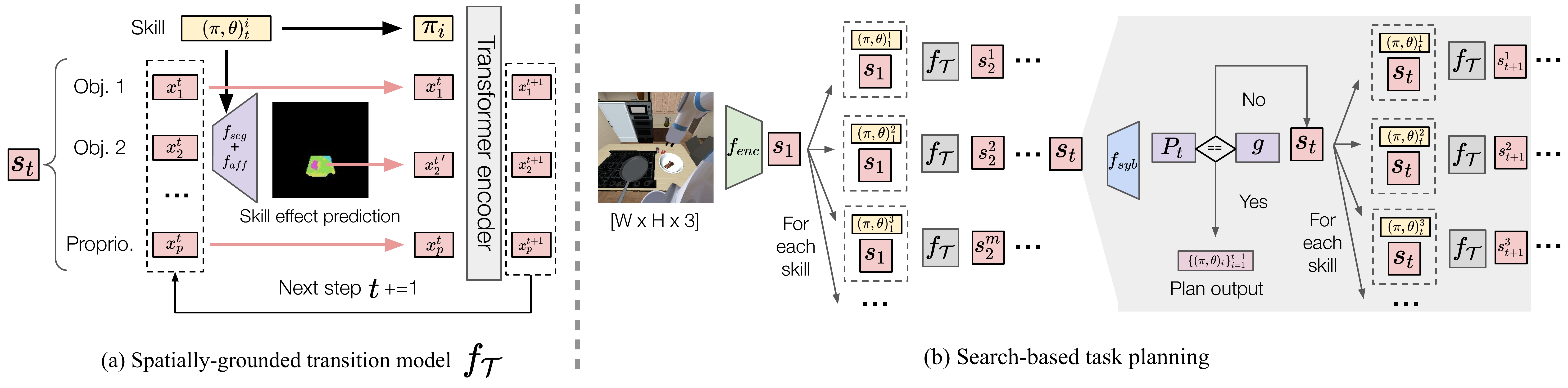}
	\caption{\textbf{Task Planning Overview}: (\textbf{a}) The spatially-grounded transition model $f_{\mathcal{T}}$ takes a set of object-level embeddings as input and generates a pixel-level prediction of the skill effect. Each pixel feature on the prediction map refers to next symbolic state of the target object after the skill execution. To handle the situation where one action affects multiple objects, a transformer encoder network takes the predicted target object feature and the other objects' state embeddings as inputs and outputs a final prediction of the skill effect over all objects. (\textbf{b}) We develop a search-based planner based on the learned transition model. For each state node, the planner enumerates possible skill and parameter choices and traverse to their effect states. Each states is then decoded into symbolic predicates and compared with a task goal. A task is considered complete if and only if all predicate conditions are satisfied.}
	\label{fig:overview}
\end{figure*}

We propose a two-stage framework to achieve generalizable task planning with parameterized skills. In the first stage, an observation encoder is trained to extract object-level features from a large-scale scene understanding dataset. The goal is to discover a suitable state representation space that could generalize to a wide variety of objects. In the second stage, we develop a novel task planner that (1) learns environment dynamics in the same representation space obtained in the first stage and (2) plans for long-horizon tasks using a search-based strategy. Here we first lay down the task planning problem setups, then describe the representation learning scheme, and finally present the learning-to-plan algorithm.

\subsection{Problem Setup}
\label{sec:problem}

\begin{figure}[!t]
	\centering
    \includegraphics[width=0.9\linewidth]{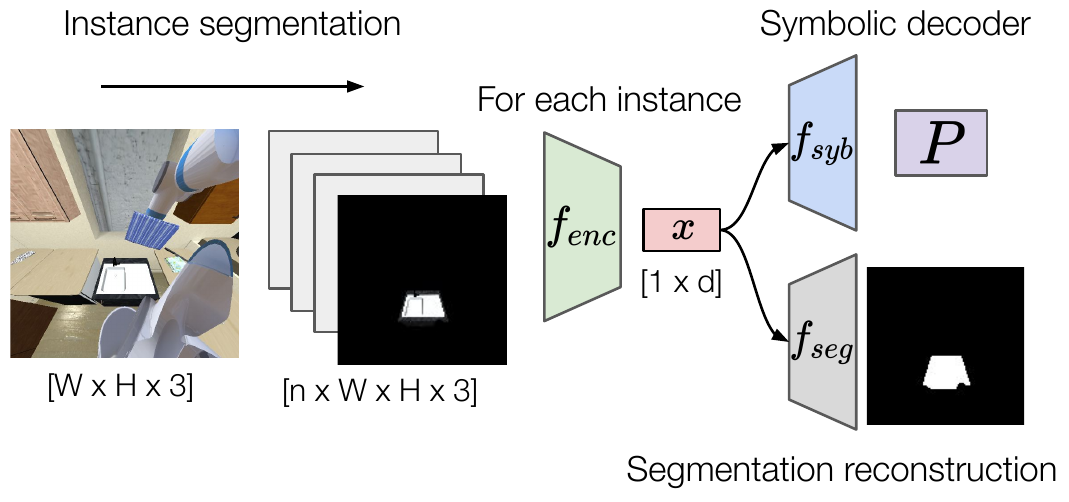}
	\caption{\textbf{Learning planning representations}: The hybrid objective consists of a high-level symbol grounding and a pixel-level segmentation reconstruction task.}
	\label{fig:pretraining}
\end{figure}

\textbf{State representation}: Consider a multi-step manipulation planning problem in a partially-observable setting. The goal is to take observation $o \in O$ as inputs and plan for a sequence of parameterized skills (described later) to reach a task goal $g \in G$. The goal space $G$ in this work is a set of known symbolic predicates $g = \{p_1, p_2, \dots \}$ that summarize the object-level state information of the environment. For example, the goal of a cooking task can be expressed as $p_1 = \texttt{Cooked(food)}$, $p_2 = \texttt{OnTop(food, plate)}$. To bridge the gap between the symbolic representation and raw observations, we adopt an object-centric observation space and state representation. We follow prior works~\cite{xu2019regression,mukherjee2020reactive,qureshi2021nerp} and assume the observation is given in the form of segmentation-masked images for each object (Fig.~\ref{fig:pretraining}). In the representation learning stage (Sec.~\ref{sec:representation}), we train an observation encoder to map the object-centric observations into an implicit representation space $f_{obs}: O \rightarrow \mathcal{S}$, where each state $s = \{x_0, x_1, \dots, x_n\}$ is a set of object-level embeddings. The representation space serves as a foundation for the subsequent learning-to-plan stage (Sec.\ref{sec:skilleffect}).

\textbf{Parameterized skills}: Following prior works \cite{ames2018learning, xu2021deep, liang2021search}, we define a parameterized skill as a policy $\pi \in \Pi$. Each skill $\pi$ is modulated by a parameter $\theta$ that specifies skill-specific execution details such as the 3D grasping poses for the grasping skill. In our task-planning formulation, there are two additional elements for a given parameterized skill: a pre-condition identifier that specifies whether the skill is \textit{executable} with the given parameters $\theta$, and a state transition model $s_{t+1} = f_\mathcal{T}(s_t, \pi, \theta)$. Different from prior task planning works which either assume the pre-conditions of the skills are given \cite{liang2021search, garrett2020pddlstream} or the transition model is pre-defined \cite{garrett2020pddlstream, curtis2021long}, our method treats both as unknown and should be learned from experience data (Sec.~\ref{sec:skilleffect}).

\textbf{Search-based task planning}: 
The task planning problem is to find a parameterized skill plan $\{(\pi, \theta)_t\}_{t=1}^T$ such that the goal condition $g$ is satisfied at the end of the last skill. Recall that goals are conjunctions of symbolic predicate conditions. A task is successful if and only if all conditions are satisfied. We adopt a breadth-first search strategy to plan for both skill skeletons and parameters (Sec.~\ref{sec:search}). 

\subsection{Representation Learning for Planning}
\label{sec:representation}

A challenge for planning with raw sensory observation is to extract suitable representation. TAMP methods rely on carefully-designed abstractions such as object poses or semantic states \cite{garrett2020pddlstream, liang2021search}. These representations are often domain-specific and require extensive hand-engineering. Other works learn state representation together with the world models in an end-to-end fashion\cite{hafner2019learning,xu2021deep}. However, these methods require interactive experience data and the learned representation often overfits to the specific task settings. Recent works in visual reasoning \cite{mao2019neuro,yuan2021sornet} showcase that it is possible to learn implicit representation from scene understanding tasks that could transfer to unseen objects. However, as we will show empirically, representations containing only abstract concepts are not sufficient for manipulation planning. Modeling the effects of parameterized skills also requires more detailed information such as spatial location and shape. In this work, we propose a suite of supervised tasks to encode information at different abstraction levels.

Specifically, we consider a high-level symbol grounding task and a pixel-level segmentation reconstruction task as shown in (Fig.~\ref{fig:pretraining}). We use an encoder-decoder structure to jointly optimize the two objectives. First, an observation model $f_{enc}$ maps the object-centric observations into object embedding space $S$. The symbol grounding task is to predict a set of high-level symbolic states $P$ for each embedding. We consider two types of predicates: object-level states (unary) and relationships between objects (binary). Each predicate has an individual decoder network $f_{syb}$ that is shared across objects. The network takes a concatenation of two object embeddings as input for binary predicates. For the segmentation reconstruction task, we use a CNN-based decoder $f_{seg}$ to reconstruct the object's segmentation mask from its feature vectors. Intuitively, this task is to instill a compact encoding of the shape and the location information in the learned object representation. To satisfy both supervised learning objectives, the $f_{enc}$ network has to extract both high-level symbolic features and low-level geometrical locations of the object from the observation. Note that, our framework is not limited to these two types of information encoding and could potentially incorporate additional supervisions to enrich the representation for different manipulation goals.

\subsection{Learning Transition Model}
\label{sec:skilleffect}

Another challenge of task planning is to predict the effects of skills so that the robot could determine the best skill plan for a task goal. To enable such capability, traditional methods rely on hardcoded analytical transition models \cite{toussaint2015logic, garrett2020pddlstream}. But this may not always be feasible, especially for complex dynamics such as nonprehensile contacts. For learning-to-plan methods, although the transition model can be learned from experience data~\cite{hafner2019learning, tian2020model, xu2021deep, chitnis2021learning}, it is often tied to the training environments and requires additional finetuning to transfer to new objects and task settings.

Here we present a learning-to-plan method that builds a generalizable transition model by leveraging the learned object representations. As described above, the pretraining scheme allows the object embeddings to focus their representation powers on task-relevant features. For example, learning to predict the relationship \texttt{on-top} encourages the encoder to discard appearance features and focus on the spatial locations of the objects. Accordingly, a transition model built within such a representation space can potentially generalize its knowledge about the skill effects to new objects. For example, the effect of a placing skill depends only on the shape of an object and the location of the supporting surface instead of their textures. 

To realize this intuition, we need to address two technical challenges: (1) How to constrain the learned transition to be within the pre-trained representation space. (2) How to ground parameterized skills on the implicit representations.

\textbf{Learning generalizable transition model.}
To allow the transition model to inherit the generalizability of the learned representation, we must constrain its prediction to the same representation space $S$. We employ two regulatory objectives: state regression and representation alignment. Given an encoded trajectory $\{(s_t, \pi_t, \theta_t)\}_{t=1}^T$, state regression is to train the transition model $f_\mathcal{T}(\cdot)$ to predict future states in the learned representation space through an L2 loss.
\begin{align}
L_{reg} = \sum_{t=1}^{T-1}||f_\mathcal{T}(s_t, \pi_t, \theta_i) - f_{enc}(s_{t+1})||
\label{equ:lossreg}
\end{align}

However, optimizing for this objective alone cannot guarantee that the predicted state is still in $S$, because a small error in the representation space may be amplified during decoding. Hence we introduce a second objective to \emph{align} the predicted representation using the same learning objectives described in Sec.~\ref{sec:representation}. More specifically, we freeze the decoder networks $f_{syb}$ and $f_{seg}$ and use the gradient from their prediction errors to update the object representations. Intuitively, these objectives encourage the transition model to predict future states that are (1) close to the reference states and (2) can be decoded into meaningful symbolic predicates and object segmentations.

\textbf{Spatially-grounded transition model.} Here we dive into the details of the transition model $f_\mathcal{T}(s, \pi, \theta)$. A major challenge of modeling the effects of a parameterized skill is to associate its parameters with its target object. For example, it is hard to predict whether a 6-D grasping pose in the robot coordinate frame will successfully grasp the handle of a pan represented as image observation. Inspired by prior works in visual skill affordance~\cite{zeng2018robotic, xu2021deep,zeng2020transporter}, we propose to \emph{ground} skill parameters \emph{spatially} onto the image observation space. 

We follow~\cite{zeng2018robotic,zeng2020transporter} and assume common manipulation skills such as grasping and placing can be parameterized by rays cast from the observation image plane. We then use depth signals to recover a location in 3D where the ray intersects with the surface of an object. We make simplifying assumptions about low-level manipulation and use this location to parameterize various skills such as, e.g., grasping location and placing pose. Thus each pixel location (a ray) can be mapped to a continuous skill parameter. We defer more complex skill parameterizations and realistic manipulation to future works. 

Having established connections between pixel space and skill parameter space, we are ready to introduce our spatially-grounded transition model. On a high level, the model learns to predict the effect of applying a skill on a target object. And it generates predictions for all parameter choices included in an image plane. As shown in Fig~\ref{fig:overview}(a), the proposed transition model has two components. First, the model decodes each object embedding $x_i$ to an object embedding map of shape $[H \times W \times d]$, with each pixel representing the effect of a skill parameterized by that pixel location. However, the prediction assumes the skill has effects confined to the target object. But most manipulation skills affect more than one object. As a simple example, a pouring skill changes both the state of the receptacle and the container in hand. Hence the second component in our transition model is a transformer encoder network \cite{vaswani2017attention} that models how applying a skill may influence each object in a scene. The transformer network takes the skill-effect prediction on the target object and all the other objects' current state embeddings as inputs, and learns to reason the dependency between objects and outputs the skil-effect prediction for each object. Together, the two components allows us to model the effects of parameterized skills as a single-step transition $s_{t+1} = f_\mathcal{T}(s_t, \pi, \theta)$. We apply the model autoregressively to predict the outcome of a multi-step plan.

\subsection{Planning with Learned Transition Models}
\label{sec:search}

\begin{algorithm2e}[!t]
\KwOut{$\mathcal{F}$ \Comment{A list of potential next states}}
\KwIn{$s_t$, $\Pi$, $f_{\mathcal{T}} = \{f_{seg}, f_{aff}, f_{trans}\}$, $f_{syb}$}
Initialize $\mathcal{F} \leftarrow \emptyset$ \\
\For{$\pi$ in $\Pi$}{
\For{$x$ in $s_t$}{
$\mathbf{M} = f_{seg}(x)$  \Comment{object segmentation mask} \\
$\mathbf{E} = f_{aff}(\pi, x)$ \Comment{skill-effect prediction map} \\
$\mathbf{E} = \mathbf{M} \cdot \mathbf{E}$ \Comment{mask out irrelevant pixels} \\
$\mathbf{S}_{t+1}= f_{trans}(s_t, \mathbf{E})$ \Comment{skill-effect prediction} \\
$\mathbf{P}_{t+1}, \mathbf{C}_{t+1} = f_{syb}(\mathbf{S}_{t+1})$ \Comment{symbolic state decoding along with confidence score} \\
\For{$P_{t+1}$ in \texttt{set}$(\mathbf{P}_{t+1})$}{
$\theta = \arg\max(\mathbf{C}_{t+1}[\mathbf{P}_{t+1} == P_{t+1}])$ \\
$\mathcal{F}$\texttt{.append}$((\pi, \theta, \mathbf{S}_{t+1}[\theta]))$
}
}
}   
return $\mathcal{F}$
\caption{Next State Sampling with Learned $f_\mathcal{T}$}
\label{alg:skilleffect}
\end{algorithm2e}

The goal of a task planner is to find a sequence of parametrized skills that are likely to reach a task goal (Sec.~\ref{sec:skilleffect}). More importantly, the planner should be able to plan for different task goals without need of re-training or finetuning. Because our transition model trained with task-agnostic exploration data, a planner can use it to search for any reachable goals that are represented in the transition model. However, the major challenge of searching for a multi-step manipulation task is to \textit{efficiently} find the plan and skill parameters in a large planning space. Since enumerating all skill parameters is infeasible, it is crucial to prune less promising branches to accelerate the search. In this section, we describe the implementation details of how we leverage the learned transition model to sample and constrain the searching branches. The sampling algorithm is summarized in pseudocode in Alg.\ref{alg:skilleffect}.

Given the state input $s_t$ at time step $t$, we first search over skill selection $\pi\in \Pi$ and target object $x \in s_t$. For each pair $(\pi, x)$, the segmentation decoder network $f_{seg}$ first localizes the target object by predicting a binary segmentation mask $\mathbf{M}$ from the object embedding $x$. The skill-effect network $f_{aff}$ takes the skill and object feature as inputs and predicts a dense pixel-wise skill effect map $\mathbf{E}$. We mask out the irrelevant pixel features using the predicted segmentation $\mathbf{M}$. To handle the situations where a skill may affect more than one object, the transformer network $f_{trans}$ takes each predicted skill-effect feature and all other objects' current state feature as inputs and generates a final pixel-wise skill effect map of the next state $\mathbf{S}_{t+1}$.

The next step is to sample states from the pixel prediction $\mathbf{S}_{t+1}$. However, searching over each pixel is time consuming and most of the pixels are redundant since they would reach the same symbolic state. We empirically find that our pre-trained symbolic decoder network $f_{syb}$ provides useful confidence signal to prune less promising paths. We infer a prediction confidence score by summing up the binary classification score of the symbolic predicates for each pixel. For each unique symbolic state $P_{t+1} \in \mathbf{P}_{t+1}$, we select the one with the highest confidence score and use its pixel location $\theta$ as the control parameters for the execution of skill $\pi$. This way, we could largely reduce the search complexity. The Appendix.\ref{app:time} includes an empirical results on search time efficiency. The bolded variables in Alg.\ref{alg:skilleffect} are pixel-wise prediction results and $\theta$ is the skill parameter.

Finally, a breadth-first search algorithm uses the state sampling sub-procedure to conduct task planning. As is illustrated in Fig.\ref{fig:overview}(b), the symbolic state $P_t$ of each sampled next state is checked against the task goal $g$. If multiple skill plans are found to reach the goal, the plan with the highest accumulated confidence score is executed. Appendix.\ref{app:details} includes more details of the search process.

\section{Experiments}

\begin{figure*}[t]
	\centering
    \includegraphics[width=1.0\linewidth]{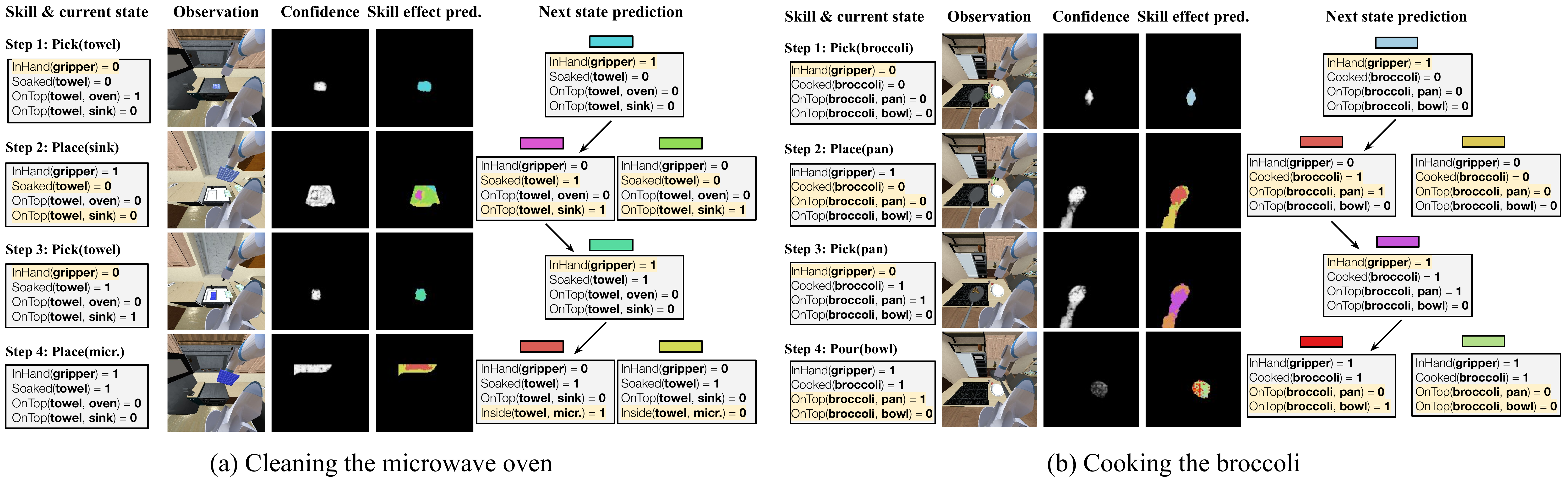}
	\caption{Qualitative results of the pixel-level skill effect prediction in the Cleaning and Cooking tasks. The leftmost column is the states prior to applying the skills. The rightmost column shows a simplified view of the tree search process. Each color on the embedding map represent the predicted effect of a skill in symbolic space. The correspondence between the pixel color and its representing symbolic state is shown on the right side of each figure.
	}
	\label{fig:exp_qualitative}
\end{figure*}

In this section, we aim to validate our hypothesis that:
(1) the object representations obtained from the pretraining stage is suitable for multi-step task planning,
(2) our spatially-grounded transition model can better model the effects of parameterized skills compared to an ungrounded model~\cite{xu2021deep}, and
(3) our learning-based task planner built on the pre-trained representations can generalize to unseen objects and improve the sample efficiency of learning to plan in different environments.

\subsection{Task setup} 

We test our models and baselines in three simulated manipulation tasks that are adapted from the BEHAVIOR benchmark\cite{srivastava2022behavior} and simulated using PyBullet\cite{coumans2016pybullet}. BEHAVIOR is a collection of long-horizon tasks that strives to capture the complexity of real-world household activities. The tasks feature a wide range of kinematic and semantic states. For example, a food item can be \texttt{OnTop} of a pan and \texttt{Cooked}. These states are both represented symbolically and externalized as visual appearance changes. 

For the \textit{Rearrangement} task, the initial state is \texttt{OnTop(A, sink)} $\land$ \texttt{OnTop(B, oven)}, the goal of the task is \texttt{OnTop(B, sink)} $\land$ \texttt{OnTop(A, oven)}. Here \texttt{A}, \texttt{B} are placeholders for objects and we use objects within the training data (\textit{Seen}) and those not included in the training set (\textit{Unseen}) to evaluate generalization. For the \textit{Cleaning} task, the initial state is \texttt{OnTop(towel, oven)} $\land$ $\lnot$\texttt{Soaked(towel)} and the goal state is \texttt{InSide(towel, microwave)} $\land$ \texttt{Soaked(towel)}. This requires the agent to first soak the towel with water in the sink then place the towel into the microwave oven for cleaning the dirt marks inside. \textit{Rearrangement} and \textit{Cleaning} share the same environment and training datasets. And we use the same learned model to plan for both tasks. For the \textit{Cooking} task, the initial state is \texttt{OnTop(A, table)} $\land$ $\lnot$\texttt{Cooked(A)} and the goal state is \texttt{OnTop(A, bowl)} $\land$ \texttt{Cooked(A)}. This requires the agent to first cook the food item \texttt{A} with a pan then serve it on a bowl.

\subsection{Parameterized skills} 

The agent is provided with three location-based parameterized motor skills: \emph{pick}, \emph{place} and \emph{pour}. \emph{Pick} executes top-down grasps parameterized by a 3D grasping location. Similarly, \emph{place} releases a grasped object at a certain 3D location relative to the agent. \emph{pour} moves a container object (e.g. pan) to a 3D location and execute a fixed pouring motion. The motion trajectory of all skills are generated using RRT-based\cite{kuffner2000rrt} motion planners.

\begin{figure}[t]
	\centering
    \includegraphics[width=0.9\linewidth]{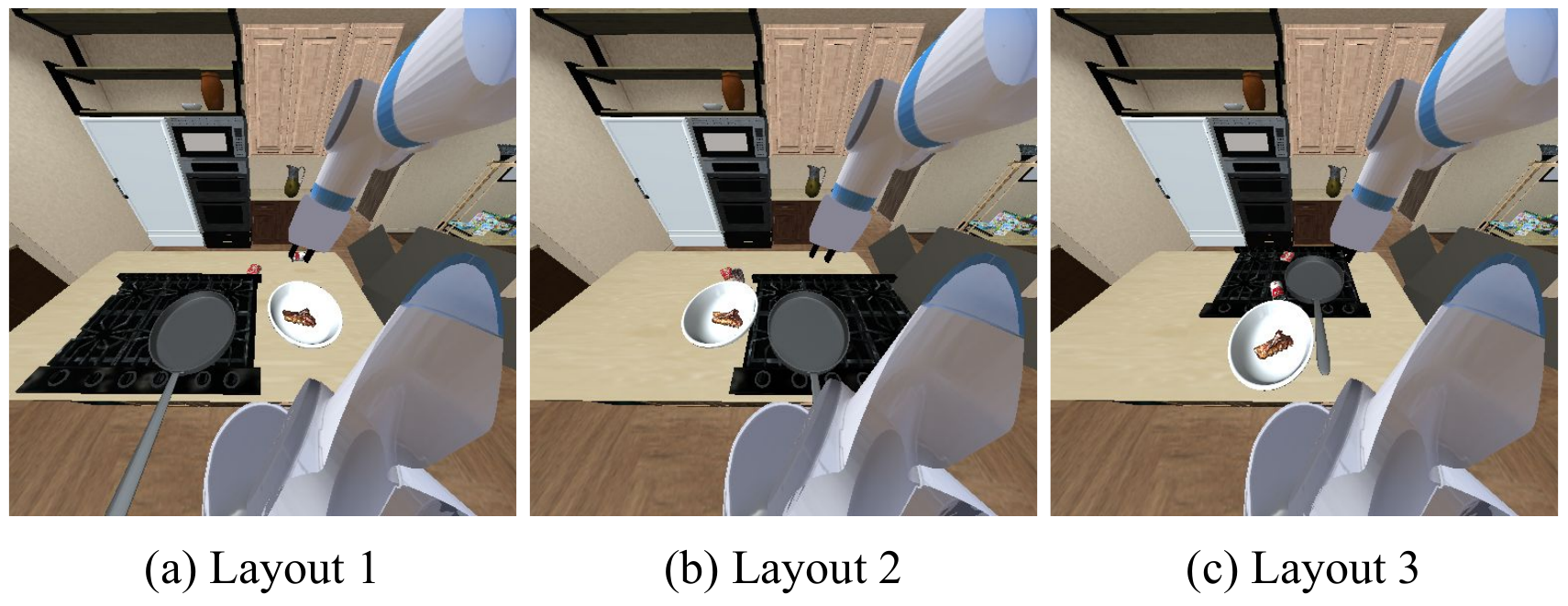}
	\caption{Three different layouts in the Cooking task.}
	\label{fig:exp_layout}
\end{figure}

\subsection{Dataset} 

We make use of two data sources. The first is a scene understanding dataset for representation learning. The content is synthesized using the BEHAVIOR engine by randomizing both the kinematic and semantic states of a wide range of objects. The second is an experience dataset for learning the state transition model $f_\mathcal{T}$, which is collected through random robot exploration. Here we describe the main factors of variations in our datasets and evaluation protocols.

\textbf{Objects} We mainly consider five object categories in our experiments: $27$ food objects, $18$ household objects, $20$ bowl containers, two cleaning towels, and a cooking pan. The food items can be set to \texttt{Cooked}. The cleaning towels can be \texttt{Soaked} indicating whether they have been soaked in the sink. The rest of the objects categories can only experience kinematic state changes. During the data collection of the representation learning dataset, we divide the first three categories into a \textit{Seen} object set used for synthesizing the representation learning dataset ($20$ food instances, $13$ household objects, $15$ bowls) and set aside the rest as \textit{Unseen} set for evaluation. In this work, we focus on exploring the model's category-level generalization, which means both \textit{Seen} and \textit{Unseen} instances are within known object categories. For the collection of the experience dataset, we only use a small subset of \textit{Seen} objects ($1$ food, $2$ household objects, $1$ bowl). The goal is to evaluate the sample efficiency and generalization capability of each method subject to a limited amount of experience data for learning environment dynamics.

\textbf{Environment layouts} For all tasks, we randomize the initial positions of task-relevant objects in confined areas. In the \textit{Cooking} task, we further create three different layouts by swapping the location of the objects (Fig.\ref{fig:exp_layout}). This creates a harder learning problem as the model has to adapt to different initial configurations.

\textbf{Exploration data} With the pre-defined parameterized skills, we let the agent explore the environment with randomly selected skill and skill parameters. The collected exploration dataset is used for training the transition model $f_{\mathcal{T}}$. To improve the diversity of the data, we apply sampling heuristics such as the same skill won't be sampled consecutively and the skill parameters are only sampled within the segmentation mask of the target object. For each environment, $500$ exploration trajectories with the sequence length of $20$ are sampled.

\subsection{Architectures and baselines}

We compare our method against three baselines: (1) Learning transition models from scratch (\textbf{Ours w.o. pre}). (2) Learning transition model without the geometric representation alignment objective (\textbf{Ours w.o. seg}). This is to demonstrate the value of geometric information in the pretrained representation. (3) A search-based variant of Deep Affordance Foresight~\cite{xu2021deep}, which can be viewed as an unstructured transition model for ungrounded parameterized skills. Here we evaluate both DAF with and without the pretrained representation (\textbf{DAF} and \textbf{DAF w.o. pre}).  We train each method with $3$ random seeds and report the average success rate ($30$ total trials for each seed) of reaching the task goal after executing the outputted plan in Tab. \ref{exp:rearrangement}, \ref{exp:cleaning}, \ref{exp:cooking}. 

\subsection{Results}
\label{sec:results}

\begin{figure}[t]
	\centering
    \includegraphics[width=0.9\linewidth]{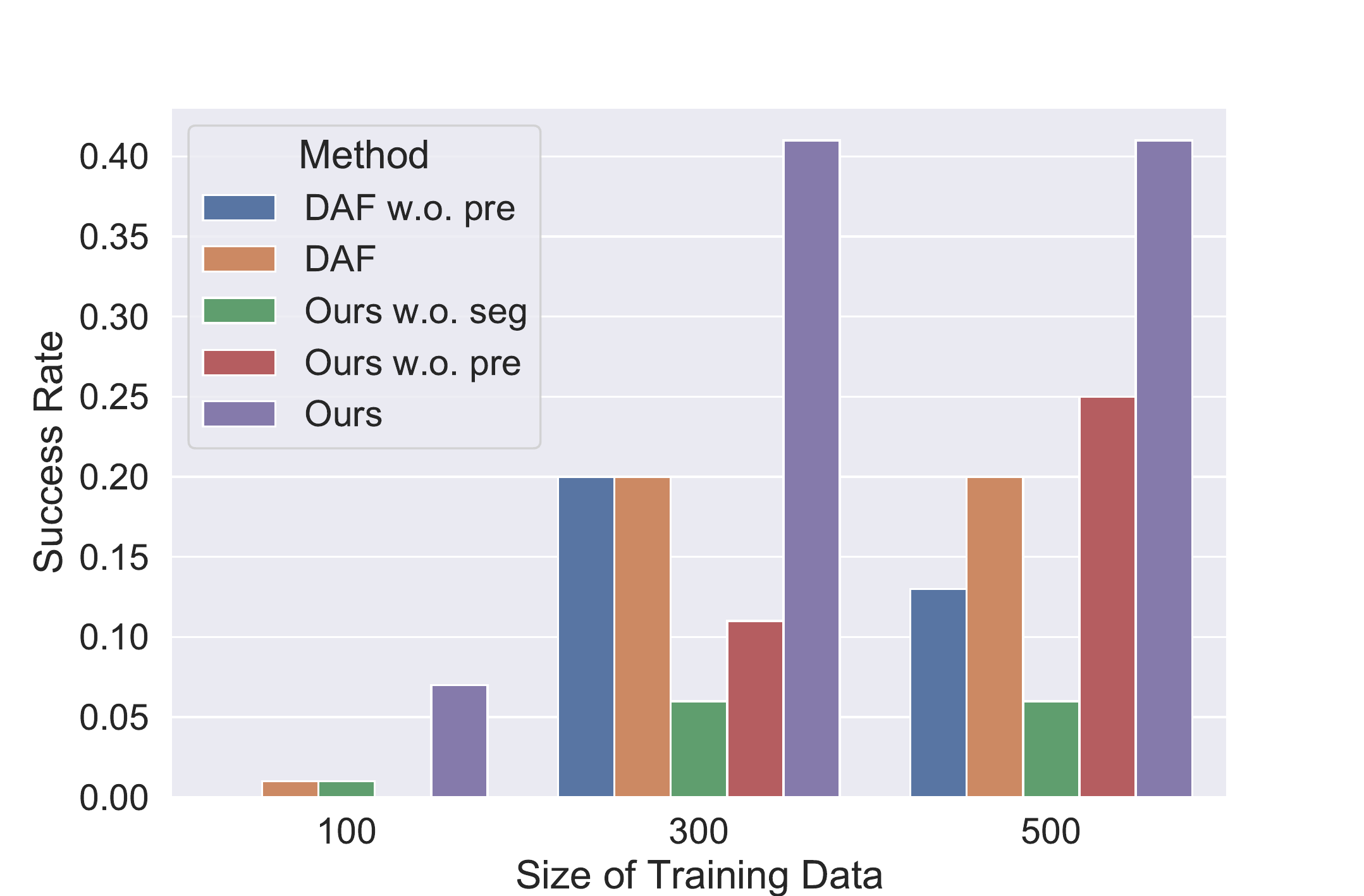}
	\caption{Evaluation results with different amount of training samples in the Cooking task with three environment layouts.}
	\label{fig:exp_layout_results}
\end{figure}

\begin{table}[]
\centering
\caption{Results of the \textbf{Rearrangement} task}
\setlength{\tabcolsep}{2mm}{
\begin{tabular}{l|ccccc}
\toprule
Switching & \begin{tabular}[c]{@{}c@{}}DAF\\ w.o. pre\end{tabular} & DAF & \begin{tabular}[c]{@{}c@{}}Ours\\ w.o. seg\end{tabular} & \begin{tabular}[c]{@{}c@{}}Ours\\ w.o. pre\end{tabular} & Ours \\ \hline
Seen & 0.65 & \textbf{0.71} & 0.29 & 0.68 & 0.65 \\
Unseen 1 & 0.01 & 0.62 & 0.31 & 0.1 & 0.47 \\
Unseen 2 & 0 & 0.34 & 0.28 & 0.02 & 0.39 \\
Unseen 3 & 0.01 & 0.26 & 0.33 & 0.01 & 0.56 \\
Mean unseen & 0.01 & 0.41 & 0.31 & 0.04 & \textbf{0.47} \\ \bottomrule
\end{tabular}
}
\label{exp:rearrangement}
\end{table}

\begin{table}[]
\centering
\caption{Results of the \textbf{Cleaning} task}
\setlength{\tabcolsep}{2mm}{
\begin{tabular}{l|ccccc}
\toprule
Cleaning & \begin{tabular}[c]{@{}c@{}}DAF\\ w.o. pre\end{tabular} & DAF & \begin{tabular}[c]{@{}c@{}}Ours\\ w.o. seg\end{tabular} & \begin{tabular}[c]{@{}c@{}}Ours\\ w.o. pre\end{tabular} & Ours \\ \hline
Seen & 0.16 & 0.08 & 0.14 & \textbf{0.44} & 0.37 \\ \bottomrule
\end{tabular}
}
\label{exp:cleaning}
\end{table}

\begin{table}[]
\centering
\caption{Results of the \textbf{Cooking} task}
\setlength{\tabcolsep}{2mm}{
\begin{tabular}{l|ccccc}
\toprule
Cooking & \begin{tabular}[c]{@{}c@{}}DAF\\ w.o. pre\end{tabular} & DAF & \begin{tabular}[c]{@{}c@{}}Ours\\ w.o. seg\end{tabular} & \begin{tabular}[c]{@{}c@{}}Ours\\ w.o. pre\end{tabular} & Ours \\ \hline
Seen & 0.09 & 0.30 & 0.24 & 0.10 & \textbf{0.48} \\
Unseen 1 & 0.09 & 0.34 & 0.22 & 0.04 & 0.32 \\
Unseen 2 & 0.04 & 0.12 & 0.13 & 0.03 & 0.30 \\
Unseen 3 & 0.03 & 0.12 & 0.16 & 0.03 & 0.31 \\
Mean unseen & 0.06 & 0.19 & 0.17 & 0.04 & \textbf{0.31} \\ \bottomrule
\end{tabular}
}
\label{exp:cooking}
\end{table}

\textbf{Hybrid semantic-geometric representation is important for manipulation planning}. In all three experiments, the model learned without low-level segmentation details (Ours w.o. seg) underperforms Ours in both \textit{Seen} and \textit{Unseen} object settings. Moreover, we observe that when training the model with data collected from different environment layouts (Fig. \ref{fig:exp_layout}), \textbf{Ours w.o. seg} failed to generate reasonable task plans. 

\textbf{Spatially-grounded transition model outperforms unstructured transition model}. While DAF achieves performances that are comparable to our method in the easiest \textit{Rearrangement} task, its performance drops significantly when planning for tasks that require more precise skill parameter selections such as putting the towel into the sink in the \textit{Cleaning} task and pouring the food into the bowl in the \textit{Cooking} task. Our approach with the spatially-grounded transition model outperforms DAF for more than $15\%$ with \textit{Seen} objects in these two environments.

\textbf{Pre-trained representation is the key for generalization to new objects}. We evaluate our approach and DAF both with and without the pre-trained representation space in unseen object settings, where we replace the towels, foods, bowls and household objects with unseen instances that not included in the scene understanding dataset and the experience dataset. In Tab.\ref{exp:rearrangement}, both models trained without the pre-trained representation (\textbf{DAF w.o. pre} and \textbf{Ours w.o. pre}) suffer a huge performance drop from $>60\%$ success rate down to close to $0\%$ results after Rearrangment the testing objects from seen to unseen. In contrast, both models with the pre-trained representation has a smaller performance drop ($20\%$ for \textbf{Ours} and $30\%$ for \textbf{DAF}). \textbf{Ours} achieves the best result in both \textit{Rearrangement} and \textit{Cooking} tasks, which outperforms the other baselines for more than $10\%$ in the most challenging cooking unseen food task.

\textbf{Pre-trained representation improves the sample-efficiency of learning in multiple environments}. In Fig.\ref{fig:exp_layout}, we observe that our approach with the pre-trained representation outperforms the baseline methods in all three training settings with different amounts of training samples. Noticeably, with only $60\%$ of the training data, \textbf{Ours} outperforms baselines for more than $20\%$ in success rate. Although \textbf{DAF} also leverage the pre-trained representation, their performance doesn't show much improvement. This also highlights the effect of our spatially-grounded transition model on generalizing to different task environments.

\textbf{Visualizing learned transitions.} We visualize the skill effect predictions from our transition model in Fig.\ref{fig:exp_qualitative}. Each colored pixel on the skill-effect prediction map shows the predicted next symbolic state when a skill is applied at that location. The confidence map is the pixel-wise symbolic prediction classification score introduced in Sec.~\ref{sec:search}. For example, the second step of the cleaning task shows placing the towel at different locations on the counter leads to different future states. And the only way to reach the \texttt{Soaked} state is to place the towel inside the sink. Similarly, in the second row of the cooking task, only the cooking surface of the pan affords the \texttt{Cooked} state of the food. We also illustrate how our model might take into account the confidence of a transition prediction. Taking pouring food into bowl as an example (last row of \textit{Cooking} in Fig.\ref{fig:exp_qualitative}): The pouring skill trajectory approaches the bowl from the left side. Our model prefers pouring locations that are close to the left half of the bowl (higher confidence) to take into account the inertia of the food items to avoid overshoot.

\section{Conclusion}
We have proposed a new learning-to-plan method that leverages object representations learned through large-scale pretraining. Through a suite of tasks inspired by everyday home activities, we show that the planner can (1) solve complex multi-step manipulation tasks and (2) generalize to new task goals and objects by inheriting invariant features from the learned representation. 

\printbibliography

\clearpage
\appendix

\subsection{Implementation Details}
\label{app:details}
This section describes the implementation details of the search-based task planning framework proposed in the main paper.

With the learned state transition model, we could now solve the task planning problem with a tree search algorithm. In this work, we adopt a vanilla breadth-first search strategy to showcase the generalization capability of the transition model that is learned based on the pre-trained representation. We defer more complex search frameworks and efficient bi-level planning approaches to future works. Alg.\ref{alg:search} shows the pseudo-code for the search-based task planning procedure. Below we walk through the key steps in the procedure.

\begin{algorithm2e}[]
\KwOut{$\{(\pi, \theta)_t\}_{t=1}^{T}$ \Comment{A task plan}}
\KwIn{$o$, $g$, $D$, $f_{enc}$, $f_{syb}$}

Initialize the priority queue $\mathcal{Q} \leftarrow \emptyset$ \\ 
Initialize the result buffer $\mathcal{R} \leftarrow \emptyset$ \\
Initialize the skill sequence $\mathcal{I} \leftarrow \emptyset$ \\
Initialize the skill sequence confidence score $c = 1$ \\
Initialize the search depth $d = 0$ \\
$s_1 = f_{enc}(o)$ \\
$\mathcal{Q}$\texttt{.push}$((s_1, d, \mathcal{I}, c))$ \\

\While{$\mathcal{Q}$ is not empty}{
$(s_t, d, \mathcal{I}, c) = \mathcal{Q}$\texttt{.pop()} \\
\If{$d > D$}{
\texttt{break\_loop()}
}
$\mathcal{F} = $ \texttt{Sample\_next\_states}$(s_t)$ \\
\For{$(\pi, \theta, s_{t+1})$ in $\mathcal{F}$}{
$\mathcal{I}$\texttt{.append}$((\pi, \theta))$ \\
$P_{t+1}, c_{t+1} = f_{syb}(s_{t+1})$ \\
\If{$P_{t+1} == g$}{
$\mathcal{R}$\texttt{.append}$((\mathcal{I}, c \cdot c_{t+1}))$ \\
}
$\mathcal{Q}$\texttt{.push}$((s_{t+1}, d+1, \mathcal{I}, c \cdot c_{t+1}))$ \\
$\mathcal{I}$\texttt{.delete}$((\pi, \theta))$
}
}
return $\underset{c}{\arg\max}(\mathcal{R})$

\caption{Search-based task planning}
\label{alg:search}
\end{algorithm2e}

Given an image observation $o$ and a symbolic goal $g$ as inputs, the planner leverages the pre-trained observation encoder $f_{enc}$ and symbolic decoder $f_{syb}$ to search for a sequence of parameterized skill to reach the goal. We also provide a maximum search depth hyperparameter $D$ to limit the boundless search space. The first step is to process image observation $o$ into a set of object-level feature representation $s_1$ with the encoder network $f_{enc}$. The resulting feature state is then added to a priority queue $\mathcal{Q}$ along with the initialized current search depth $d$, confidence score $c$ and history skill sequence $\mathcal{I}$. In each iteration, the priority queue pops out a set of leaf node states. The algorithm would exit the searching phase if the current search depth exceeds the maximum search depth. If the search continues, we then use the forward sampling process as introduced in Alg.\ref{alg:skilleffect} to get a set of next state predictions. Each future state candidate are added to the priority queue with the accumulated skill sequence and confidence score. If the state prediction reaches the task goal $g$, we add the skill sequence into a result buffer $\mathcal{R}$ and return the skill sequence that has the highest accumulated confidence score as the found task plan.

One of the major challenges of a vanilla searching framework is the time efficiency, especially that we want to search over both skills and parameters sequences. The unique design of our spatially-grounded transition mode can speed up the transition model and prune the less promising paths. First, the skill transition model takes the form of a fully convolutional network, with each pixel predicting the effect of a parameterized skill, the planner can get all the prediction results with a single forward pass through the network. Second, we can treat the raw output of the symbolic state prediction as a form of confidence score and find the 3D location of the pixel that has the highest confidence as the control parameters to execute the skill. This way, we could downgrade the searching complexity to state-wise because, for each unique object state, only one pixel that has the highest confidence score is added to the search tree. Downgrading the pixel-wise searching into state-wise is the key to improving our search speed (Appendix.\ref{app:time}) and, more importantly, the classification confidence score is a byproduct of the learning process that does not require additional training or heuristics. We defer more complex pruning techniques to future works since it is not the main focus of our generalizable planning representation.

\subsection{Search time breakdown}
\label{app:time}

We record the search time cost of the vanilla pixel-wise searching and our pruned searching framework. The results are shown in Tab.\ref{exp:time}. During the evaluation, we assign task goals that require different number of planning steps (from $1$ to $5$) in the \textit{Rearrangement} task and use the same learned model to search for the task plans. Our approach is much more efficient than pixel-wise searching. That is because we leverage the predicted segmentation mask and classification confidence score to prune out the redundant pixels (Alg.\ref{alg:skilleffect}), which largely decreases the search space. For each task goal, we test $20$ rollouts and report the mean and $0.95$ confidence interval in Tab.\ref{exp:time}. For the searching that cost time longer than $1000$ seconds, we directly report the search time as $>1000$.

\begin{table}[H]
\centering
\caption{Searching time cost}
\setlength{\tabcolsep}{2mm}{
\begin{tabular}{l|c|c|c|c|c}
\toprule
 & \multicolumn{5}{c}{Searching time breakdown (seconds)} \\ \hline
\multicolumn{1}{c|}{Plan steps} & \multicolumn{1}{c|}{1} & \multicolumn{1}{c|}{2} & \multicolumn{1}{c|}{3} & \multicolumn{1}{c|}{4} & \multicolumn{1}{c}{5} \\ \hline
Pixel-wise         & 23.9$\pm$6.0          & $>1000$                       &  $>1000$                &   $>1000$               &   $>1000$          \\
Ours               & 0.9$\pm$0.1           & 2.6$\pm$0.2           & 8.9$\pm$0.8           & 21.2$\pm$2.4          & 79.6$\pm$6.1     \\ \bottomrule
\end{tabular}
}
\label{exp:time}
\end{table}

\end{document}